\documentclass[conference]{IEEEtran}
\usepackage{spconf}
\usepackage{cite}
\usepackage{amsmath,amssymb,amsfonts}
\usepackage{algorithmic}
\usepackage{graphicx}
\usepackage{textcomp}
\usepackage{tabularx,booktabs}
\usepackage{xcolor}
\usepackage{float}
\def\BibTeX{{\rm B\kern-.05em{\sc i\kern-.025em b}\kern-.08em
    T\kern-.1667em\lower.7ex\hbox{E}\kern-.125emX}}

\usepackage{caption}
\usepackage{subfig}
\usepackage{hyperref}

\begin{document}

\title{Explainability of Deep Learning models for Urban Space perception}

\name{Ruben Sangers \qquad Jan van Gemert \qquad Sander van Cranenburgh}
\address{Delft University of Technology}

\maketitle

\begin{abstract}
Deep learning based computer vision models are increasingly used by urban planners to support decision making for shaping urban environments. Such models predict how people perceive the urban environment quality in terms of e.g. its safety or beauty. However, the blackbox nature of deep learning models hampers urban planners to understand what landscape objects contribute to a particularly high quality or low quality urban space perception. This study investigates how computer vision models can be used to extract relevant policy information about peoples' perception of the urban space. To do so, we train two widely used computer vision architectures; a Convolutional Neural Network and a transformer, and apply GradCAM -a well-known ex-post explainable AI technique- to highlight the image regions important for the model's prediction. Using these GradCAM visualizations, we manually annotate the objects relevant to the models' perception predictions. As a result, we are able to discover new objects that are not represented in present object detection models used for annotation in previous studies. Moreover, our methodological results suggest that transformer architectures are better suited to be used in combination with GradCAM techniques. Code is available on Github\footnote{\href{https://github.com/rsangers/explainable\_perception}{https://github.com/rsangers/explainable\_perception}}.
\end{abstract}

\begin{IEEEkeywords}
Explainability, Deep Learning, Convolutional Neural Network, Transformer, Urban Space perception
\end{IEEEkeywords}

\section{Introduction}
In urban planning, computer vision models are increasingly being used to create maps of cities showing what areas are perceived as being safe/unsafe, beautiful/ugly, and lively/boring, see e.g.~\cite{dubey2016deep,rossetti2019explaining}. Such computer vision models are trained on survey data in which respondents are asked to score the safety and beauty level for a series of streetview images presented to them. The results of these studies provide urban planners and policy makers with important insights about areas that need policy attention, e.g. because perceived safety levels are deemed too low. 

Current computer vision models to predict urban space perceptions, however, provide little guidance for urban planners to improve the urban environment. In particular, due to the blackbox nature of deep learning based computer vision models, such models do not directly reveal what the relevant landscape objects are, based on which the model geared its predictions. In light of this, some recent studies, e.g. \cite{ramirez2021measuring}, use computer vision models to extract objects to use these in turn in readily (ex-ante) explainable regression models. However, this leads to a constrained abstraction of the image as it does not include all objects present. Important objects might therefore go unnoticed and the predictive performance might degrade.      

It is increasingly realised that being able to explain and interpret the inner workings of the models is indispensable~\cite{rudin2019stop}. Two types of explainability are distinguished in the literature: ex-ante ('before the event') and ex-post ('after the event') explainability \cite{sokol2020explainability}. Ex-ante explainability aims to specifically incorporate explainable modules into the model; ex-post explainability aims to interpret models without constraining their design. An important drawback of the former over the latter is that the explainable modules are often fairly constrained representations, leading to high model biases and thereby degrading the model performance\cite{vidal2021deep}.

This study aims to shed light on how computer vision models can be used to extract relevant policy information about peoples' perception of the urban space, without relying on ex-ante explainability. To do so, we train two widely used computer vision architectures; a ResNet50 Convolutional Neural Network \cite{krizhevsky2012imagenet} (CNN) and a Data Efficient Image Transformer (DeiT) \cite{touvron2021training}. As a data set we use  a widely used used urban perception set \cite{dubey2016deep}. For explainability, we apply a GradCAM \cite{selvaraju2017grad} -a successful ex-post explainable AI technique- to highlight what part of an image is important for the model's prediction. GradCAM exploits the gradients of the last convolutional layer to create heatmaps of image pixels that were the most important for its prediction \cite{selvaraju2017grad}. Using the heatmaps provided by the GradCAM, we manually annotate and keep score of the objects relevant to the models' safety and beauty predictions for the 50 highest and lowest scoring images.

The first contribution of our paper is that we extract landscape objects explaining urban perceptions using an ex-post explainability technique, without relying on additional object detection or segmentation modules. To the best of the authors' knowledge, this not been tried before. Hitherto studies using computer vision to understand urban space perceptions use ex-ante explainability techniques either in their model design \cite{ramirez2021measuring}\cite{yao2019human}\cite{zhang2018measuring} or in their analysis \cite{shi2019deep}. As our approach does not rely on object detection models, we are able to discover new landscape objects that are not represented in present object detection models. This will thereby give urban planners a more complete overview of the landscape objects that affect perception.

The second contribution of this study is that we test and compare the suitability of two widely used computer vision architectures to explain urban perceptions. The first architecture we study is a conventional Convolutional Neural Network (CNN)\cite{krizhevsky2012imagenet} and the second computer vision architecture we study is the transformer\cite{dosovitskiy2020image}. Transformers are based on self-attention instead of convolutions. A series of recent papers suggest that CNN models over rely on local features \cite{brendel2019approximating}, such as textures \cite{geirhos2018imagenet}, while transformers do better in capturing global relationships. By assessing both architectures vis-a-vis we hope to shed light on which architecture is best suited for predicting and explaining urban space perception.

\section{Methods}
\subsection{Place Pulse 2.0 dataset and preprocessing}
\begin{figure}[ht!]
	\includegraphics[width=0.5\textwidth]{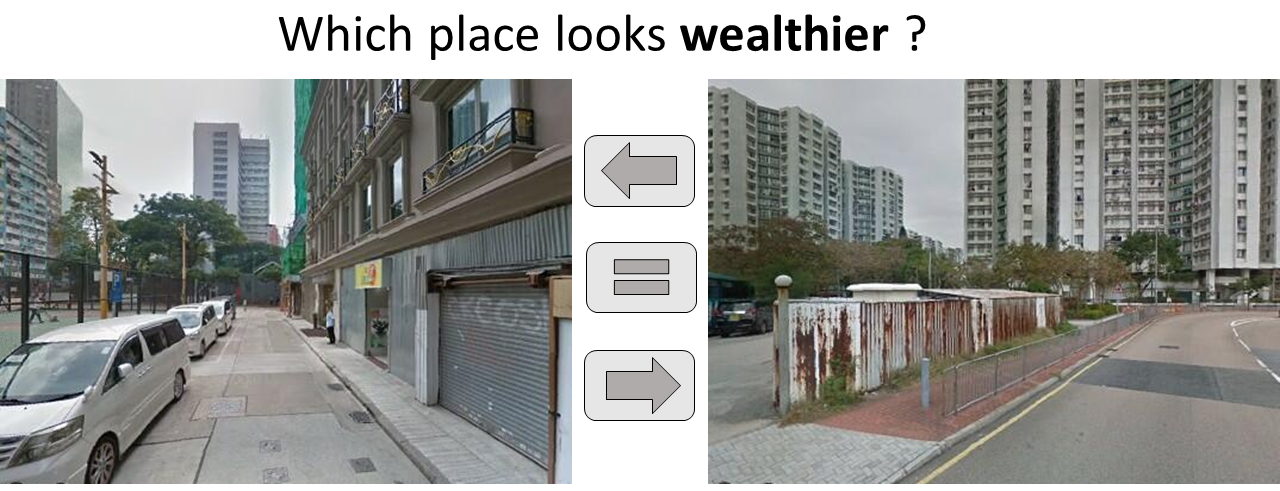}
	\caption{An example of the urban space perception assessment task. Participants are given pairwise image comparisons for a perceptual attribute, such as safety, wealthiness and depressiveness.}
	\label{fig:task_example}
\end{figure}

For the analyses of this study we use the Place Pulse 2.0 dataset \cite{dubey2016deep}. The Place Pulse dataset consists of more than 1.5 million pairwise comparisons for over 100,000 images in 56 cities, collected using the Google Street View Image API. The data are collected using crowd-sourcing. Each participant is shown a pair of images and asked to choose which one of the two images corresponds most with a queried perceptual attribute, which could be safety, liveliness, wealthiness, boringness, depressiveness and beauty (Figure \ref{fig:task_example}). In this study, we particularly focus on three the perceptual attributes, namely safety, wealthiness and depressiveness.

\subsection{Problem formulation}
The problem of predicting a perceptual attribute is mathematically defined as follows: we have an image set $I=\{x_i\}^{m}_{i=1} \in \mathbb{R}^n$ containing $m$ images, and a set consisting of $N$ pairwise comparison tuples $P=\{(i_k, j_k, y_k)\}^{N}_{k=1}$, $i,j \in \{1,...,m\}$, $y \in \{-1, +1\}$, where -1 means that image $j_k$ is preferred, while +1 denotes that image $i_k$ is preferred. We want the model to predict an attribute score $f(x_i) \in \mathbb{R}$ for an image $x_i$ such that the amount of correct image comparisons is maximized, formulated as 
\begin{equation}
    y \cdot (f(x_i) - f(x_j)) > 0, \quad \forall {(i,j,y) \in P}.
\end{equation}

To be able to approximate a solution for this problem, we need a loss function that punishes the model whenever an wrong prediction is made. Therefore, in line with the work by Dubey \cite{dubey2016deep} and Chapelle \cite{chapelle2010efficient}  we use a ranking loss. Ranking loss penalizes incorrect comparison scores by their difference, defined as: 

\begin{equation}
    L(x_i, x_j, y) = \max(0, -y \cdot (f(x_i)-f(x_j))).
\end{equation}

\subsection{Explainability study}
\textit{Training details.} We compare two current state-of-the-art networks, which are the ResNet50 \cite{he2016deep} and the Data-efficient image Transformer (DeiT) \cite{touvron2021training}. To keep training time within feasible bounds, ImageNet-pretrained versions of these networks were used, after which a transfer learning procedure was applied with a small learning rate ($\mu=10^{-3}$) for 20 epochs on the Place Pulse 2.0 dataset. 

\textit{Explainability models}
To study the ability of these computer vision architectures to explain urban perceptions, we exploit the heatmaps of GradCAM~\cite{selvaraju2017grad}. We also tested other explainable AI techniques, including: GradCAM\cite{selvaraju2017grad}, Eigen-CAM\cite{muhammad2021eigen}, Ablation-CAM\cite{ramaswamy2020ablation} and Attention Rollout\cite{abnar2020quantifying} (only for the transformer model). Based on initial results, we choose GradCAM for further analysis as it empirically showed -for both architectures- to give rise to highlighted regions that most closely overlap with the objects.

\textit{Annotating the heatmaps}
Using the heatmaps provided by the GradCAM, we manually annotate and keep score of the objects relevant to the models' safety, wealthy and depression predictions for the 50 highest and lowest scoring images. We do this without a predefined list of annotatable objects, as to not limit ourselves to the currently know object relations. Moreover, we only count an object once per image.

\section{Experiments and discussion}
\subsection{Object analysis}
Table \ref{fig:cnn_obj_table} and \ref{fig:trans_obj_table} show the main results of our study. They list the objects identified as being important for the models' predictions for the CNN based model and the Transformer respectively. Below we highlight the most important results for the CNN model and the transformer:

\vspace{0.2cm}
\textit{CNN model}
\begin{itemize}
    \item Overhanging cables show a strong association with a high perception of depressiveness and a low perception of safety and wealthiness.
    \item Trees are associated with a low perception of depressiveness and a high perception of safety and wealthiness.
    \item Building elements, such as roofs, blinds and corrugated sheets are occasionally found to be of importance for the CNN model's predictions. This is noteworthy as most previous studies consider buildings as a whole. 
\end{itemize}

\textit{Transformer model}
\begin{itemize}
    \item Tunnel roofs are often associated with a high perception of depressiveness and a low perception of safety.
    \item Regarding perceived wealthiness, 'air' is found to be of particular importance for the model's predictions.
    \item Trees are associated with a high perception of safety and a low perception of depressiveness.\\
\end{itemize}

\begin{table*}[!ht]
\caption[font=tiny]{Annotation of the objects highlighted by GradCAM in the 50 highest and lowest ranking images of each category for the CNN model. The most interesting correlations, such as the importance of power cables and vegetation, are shown in bold. The signs + and - denote images with a respectively high or low score for that perceptual attribute.}
\label{fig:cnn_obj_table}
\resizebox{\textwidth}{!}{
\begin{tabular}{|lrll|lllr|lrlr|}
\hline
\multicolumn{4}{|c|}{\textbf{Depressing}} & \multicolumn{4}{c|}{\textbf{Safety}} & \multicolumn{4}{c|}{\textbf{Wealthy}} \\ 
\multicolumn{2}{|c}{\textbf{+}} & \multicolumn{2}{c|}{\textbf{-}} & \multicolumn{2}{c}{\textbf{+}} & \multicolumn{2}{c|}{\textbf{-}} & \multicolumn{2}{c}{\textbf{+}} & \multicolumn{2}{c|}{\textbf{-}} \\ \hline
\textbf{Power cable} & \multicolumn{1}{r|}{\textbf{18}} & \textbf{Tree} & \multicolumn{1}{r|}{\textbf{16}} & \textbf{Tree} & \multicolumn{1}{r|}{\textbf{30}} & \textbf{Cable} & \textbf{21} & \textbf{Grass} & \multicolumn{1}{r|}{\textbf{14}} & \textbf{Roof} & \textbf{17} \\
\textbf{Blinds} & \multicolumn{1}{r|}{\textbf{4}} & Road & \multicolumn{1}{r|}{9} & Car & \multicolumn{1}{r|}{7} & \textbf{Roof} & \textbf{17} & Road & \multicolumn{1}{r|}{14} & Sidewalk & 10 \\
\textbf{Corrugated sheets} & \multicolumn{1}{r|}{\textbf{4}} & Car & \multicolumn{1}{r|}{8} & Sidewalk & \multicolumn{1}{r|}{4} & Wall & 7 & \textbf{Tree} & \multicolumn{1}{r|}{\textbf{10}} & \textbf{Cable} & \textbf{10} \\
Window & \multicolumn{1}{r|}{2} & Bush & \multicolumn{1}{r|}{3} & Roof & \multicolumn{1}{r|}{4} & Sidewalk & 2 & Fence & \multicolumn{1}{r|}{8} & Wall & 7 \\
Gutter & \multicolumn{1}{r|}{2} & Grass & \multicolumn{1}{r|}{3} & Road & \multicolumn{1}{r|}{3} & Car & 2 & Bush & \multicolumn{1}{r|}{6} & Tree & 4 \\
Curb & \multicolumn{1}{r|}{2} & Sidewalk & \multicolumn{1}{r|}{2} & Grass & \multicolumn{1}{r|}{3} & Door & 2 & Car & \multicolumn{1}{r|}{2} & Car & 4 \\
Sidewalk & \multicolumn{1}{r|}{2} & Stairs & \multicolumn{1}{r|}{2} & Bush & \multicolumn{1}{r|}{3} & Flat & 1 & Logo & \multicolumn{1}{r|}{1} & Fence & 3 \\
Fence & \multicolumn{1}{r|}{2} &  &  &  & \multicolumn{1}{l|}{} & Bush & 1 & Flat & \multicolumn{1}{r|}{1} & Grass & 2 \\
Car & \multicolumn{1}{r|}{1} &  &  &  & \multicolumn{1}{l|}{} &  & \multicolumn{1}{l|}{} &  & \multicolumn{1}{l|}{} & Bush & 1 \\
Clouds & \multicolumn{1}{r|}{1} &  &  &  & \multicolumn{1}{l|}{} &  & \multicolumn{1}{l|}{} &  & \multicolumn{1}{l|}{} &  & \multicolumn{1}{l|}{} \\
Pole & \multicolumn{1}{r|}{1} &  &  &  & \multicolumn{1}{l|}{} &  & \multicolumn{1}{l|}{} &  & \multicolumn{1}{l|}{} &  & \multicolumn{1}{l|}{} \\
Door & \multicolumn{1}{r|}{1} &  &  &  & \multicolumn{1}{l|}{} &  & \multicolumn{1}{l|}{} &  & \multicolumn{1}{l|}{} &  & \multicolumn{1}{l|}{} \\
Rails & \multicolumn{1}{r|}{1} &  &  &  & \multicolumn{1}{l|}{} &  & \multicolumn{1}{l|}{} &  & \multicolumn{1}{l|}{} &  & \multicolumn{1}{l|}{} \\
Bus stop & \multicolumn{1}{r|}{1} &  &  &  & \multicolumn{1}{l|}{} &  & \multicolumn{1}{l|}{} &  & \multicolumn{1}{l|}{} &  & \multicolumn{1}{l|}{} \\
Overpass & \multicolumn{1}{r|}{1} &  &  &  & \multicolumn{1}{l|}{} &  & \multicolumn{1}{l|}{} &  & \multicolumn{1}{l|}{} &  & \multicolumn{1}{l|}{} \\ \hline
\end{tabular}
}
\end{table*}
\begin{table*}[!ht]
\caption[font=tiny]{Annotation of the objects highlighted by GradCAM in the 50 highest and lowest ranking images of each category for the transformer model. The signs + and - denote images with a respectively high or low score for that perceptual attribute. Interestingly, tunnel roofs appear to have a high influence on the decision making of the model for depressiveness and safety, while wealthiness is highly determined by the sky.}
\label{fig:trans_obj_table}
\resizebox{\textwidth}{!}{
\begin{tabular}{|lrlr|lrlr|lrlr|}
\hline
\multicolumn{4}{|c|}{\textbf{Depressing}} & \multicolumn{4}{c|}{\textbf{Safety}} & \multicolumn{4}{c|}{\textbf{Wealthy}} \\ 
\multicolumn{2}{|c}{\textbf{+}} & \multicolumn{2}{c|}{\textbf{-}} & \multicolumn{2}{c}{\textbf{+}} & \multicolumn{2}{c|}{\textbf{-}} & \multicolumn{2}{c}{\textbf{+}} & \multicolumn{2}{c|}{\textbf{-}} \\ \hline
\textbf{Tunnel roof} & \multicolumn{1}{r|}{\textbf{20}} & \textbf{Tree} & \textbf{42} & \textbf{Tree} & \multicolumn{1}{r|}{\textbf{41}} & Tree & 18 & \textbf{Sky} & \multicolumn{1}{r|}{\textbf{49}} & \textbf{Sky} & \textbf{21} \\
Bush & \multicolumn{1}{r|}{5} & Road & 6 & \textbf{Car} & \multicolumn{1}{r|}{\textbf{18}} & \textbf{Tunnel roof} & \textbf{16} & Road & \multicolumn{1}{r|}{8} & Sidewalk & 10 \\
Road & \multicolumn{1}{r|}{4} & Bush & 5 & Bush & \multicolumn{1}{r|}{8} & Bush & 4 & Sidewalk & \multicolumn{1}{r|}{4} & Road & 5 \\
Window & \multicolumn{1}{r|}{3} & Car & 4 & Road & \multicolumn{1}{r|}{6} & Road & 3 & Car & \multicolumn{1}{r|}{1} & Wall & 3 \\
Tree & \multicolumn{1}{r|}{3} & Viaduct & 3 & Hedge & \multicolumn{1}{r|}{4} & Car & 3 &  & \multicolumn{1}{r|}{} & Window & 2 \\
Water & \multicolumn{1}{r|}{2} & Sidewalk & 2 & Rails & \multicolumn{1}{r|}{2} & Grass & 2 &  & \multicolumn{1}{r|}{} & Lantern & 1 \\
Wall & \multicolumn{1}{r|}{2} & Gate & 1 & Flat & \multicolumn{1}{r|}{2} & Gate & 2 &  & \multicolumn{1}{r|}{} &  &  \\
Car & \multicolumn{1}{r|}{1} & Grass & 1 & Trash container & \multicolumn{1}{r|}{2} & Window & 1 &  & \multicolumn{1}{r|}{} &  &  \\
Gate & \multicolumn{1}{r|}{1} &  & \multicolumn{1}{l|}{} & Grass & \multicolumn{1}{r|}{2} &  & \multicolumn{1}{l|}{} &  & \multicolumn{1}{l|}{} &  &  \\
 & \multicolumn{1}{r|}{} &  & \multicolumn{1}{l|}{} & Sign & \multicolumn{1}{r|}{1} &  & \multicolumn{1}{l|}{} &  & \multicolumn{1}{l|}{} &  & \multicolumn{1}{l|}{} \\
 & \multicolumn{1}{r|}{} &  & \multicolumn{1}{l|}{} & Shop & \multicolumn{1}{r|}{1} &  & \multicolumn{1}{l|}{} &  & \multicolumn{1}{l|}{} &  & \multicolumn{1}{l|}{} \\
 & \multicolumn{1}{r|}{} &  & \multicolumn{1}{l|}{} & Window & \multicolumn{1}{r|}{1} &  & \multicolumn{1}{l|}{} &  & \multicolumn{1}{l|}{} &  & \multicolumn{1}{l|}{} \\ \hline
\end{tabular}
}
\end{table*}

\begin{figure}[!ht]
	\centering
	\includegraphics[width=.45\textwidth]{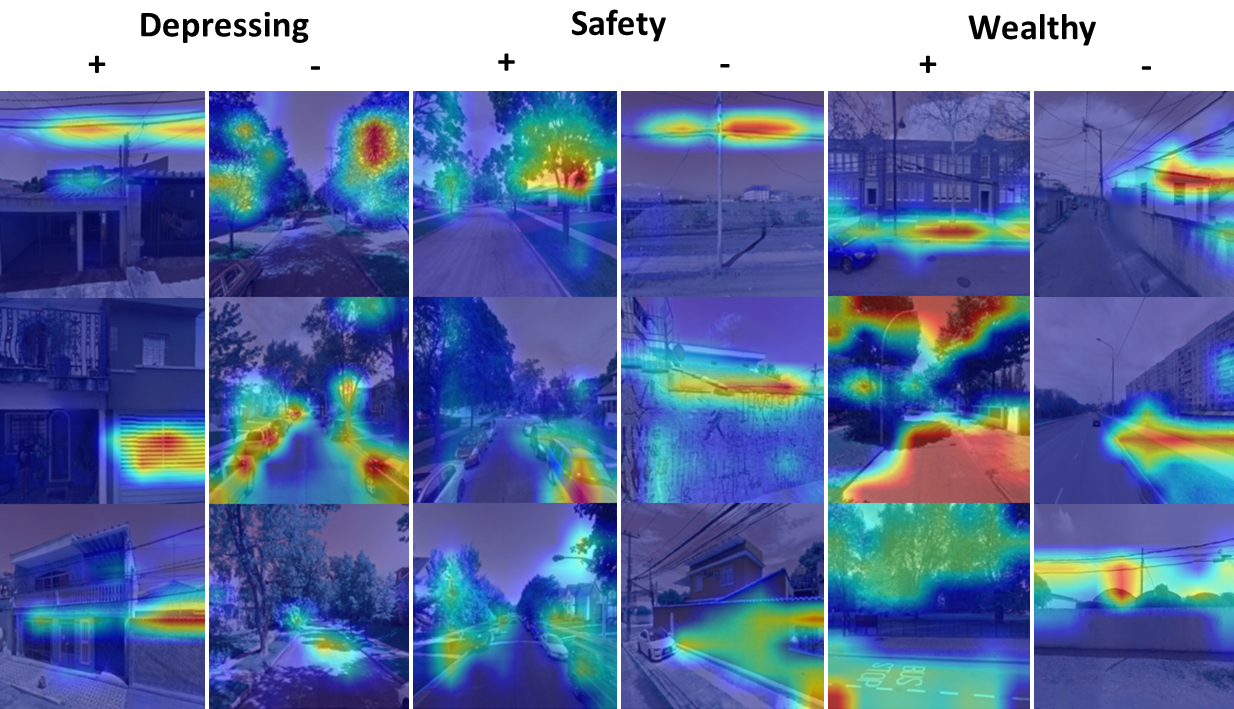}
	\caption[font=tiny]{Example images of the three most important objects for each category for the CNN model. The highlighted areas for this model are often small and not well overlapping with objects.}
	\label{fig:cnn_obj_ex}
\end{figure}
\begin{figure}[!ht]
	\centering
	\includegraphics[width=.45\textwidth]{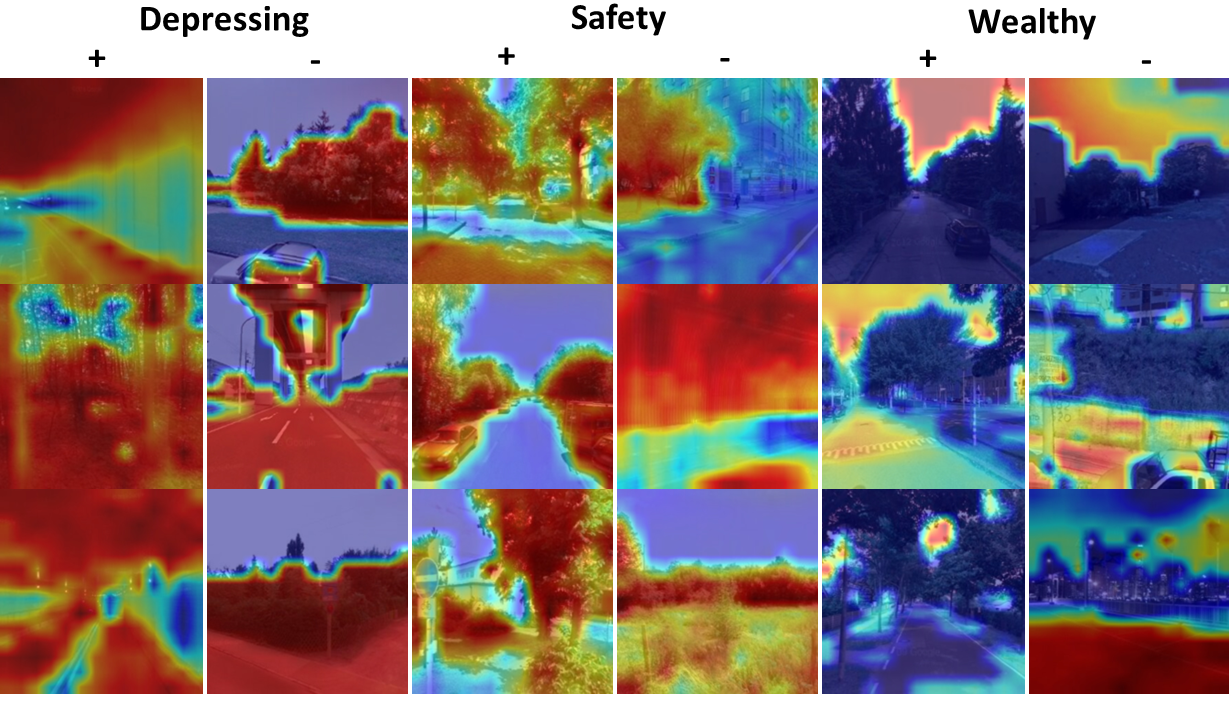}
	\caption[font=tiny]{Example images of the three most important objects for each category for the transformer model. The important regions for this model generally take up a large part of the image, and overlap relatively well with the objects in an image.}
	\label{fig:trans_obj_ex}
\end{figure}

\subsection{Explainability patterns}
Figures \ref{fig:cnn_obj_ex} and \ref{fig:trans_obj_ex} show examples of images in which the three most often encountered objects relevant for the models' prediction are present. Figure \ref{fig:cnn_obj_ex} shows images with GradCAM heatmap for the CNN model; Figure \ref{fig:trans_obj_ex} shows the same for the transformer. Comparing the GradCAM overlays across the two models, reveals that the sizes of the highlighted regions in the images are dissimilar. It can be seen that CNN model often attributes importance to a few relatively small areas in the image, while the transformer often attributes importance to a large area. Moreover, the highlighted areas of the transformer model often closely overlapped with objects present in the image, while this is much less the case for the CNN model. This finding is in line with recent literature, e.g. \cite{brendel2019approximating}\cite{geirhos2018imagenet}. 

\subsection{Discussion}
Our methodology also has some disadvantages. Firstly, it is important to keep in mind that results presented in this paper depend on a manual analysis of only 100 images per perceptual attribute and only one annotator. Especially because the explainability of images often comprise highlighted regions that do not always clearly overlap with objects, possible biases in the analysis may be present. To establish our findings more firmly, a larger scale study with multiple annotators is recommended.

Secondly, our substantive findings may be biased towards objects that are present in street imagery, such as cars, roads and trees. Objects that often appear in street imagery have an a priori higher probability to be annotated and identified as important for the model's perception prediction than  objects that appear less frequently. Hence, objects that are important for urban perceptions, but are rare to find in street imagery may go undetected in our analysis. \\

Moreover, our research may not generalise towards other models or applications. In this study we considered only two computer vision models: ResNet50 and DeiT. As such, we need to be cautious to draw strong conclusions regarding the differences between the two model architectures. Noteworthy in this regard is also the work of Shi et al.\cite{shi2019deep} and Viering et al. \cite{viering2019manipulate}. Shi et al. show that different architectures may generate different explanations. Viering et al. even show that it is possible to manipulate this sort of computer vision models in such a way that they generate any desirable explainability output generated by GradCAM, without degrading the accuracy of the model.  

Finally, the landscape objects and attributes that are important for a computer vision model to predict urban space perception, do not necessarily coincide with the objects, attributes and what more, that humans sense to perceive their surroundings. Ultimately, our models only establish associations between objects visible in images and perception scores, they do not establish causality. As such, one should act cautious when using  results of our, and similar studies, for urban planning decisions. For instance, there may be objects important to human perceptions that go undetected in street imagery but happen to correlate with objects that are detected. In that case, the policy information extracted from the model may be misguided. More concretely, the presence of small birds, like tomtits, may positively associate with human's perception of wealthiness of a place, but they may go unnoticed due to their small size in computer vision based studies, like ours. In case tomtits avoid open urban spaces, a computer vision model will (correctly) learn that open urban spaces are associated with a low perception of wealthiness. But, reducing the number of open urban space, may not be sufficient to improve perceived wealthiness, without policies to support bird wildlife.

\section{Conclusion}
This study has shed light on how computer vision models can be used to extract relevant policy information about peoples' perception of the urban space, without relying on ex-ante explainability. To gain a better understanding of the objects that are associated with human's perception of the urban space, computer vision models were trained to predict perception based on street imagery. By subsequently analyzing the decision making of these models using GradCAM, we have been able to list the objects that are the most important for the predictions of perceived safety, wealthiness and depressiveness. Some of the discovered objects are well known and reported in earlier studies, such as trees and cars, while other objects are new, such as power cables, blinds and corrugated sheets. The fact that new objects have been detected suggests that our approach can be useful for future studies with a similar substantive objective.

In this study we have considered two different computer vision architectures: a CNN and a transformer. We find that objects that are important for the models' predictions are different. In particular, the highlighted areas by the GradCAM turned out to be better interpretable for the transformer than for the CNN model, The highlighted regions for the transformer are often larger and better shaped around visible objects. We believe further research is needed to more deeply understand the differences between the explainability of the two studies computer vision architectures. Relatedly, we believe it would be beneficial to have human participants (as opposed to the machine learning researcher)  assess the degree of explainability provided by CNN and  transformer models.

\section*{Acknowledgement}
This work is supported by the TU Delft AI Labs programme. We would like to thank Tomás Ramírez and Hans Löbel for sharing their code and data with us.

\bibliographystyle{plain}
\bibliography{report.bib}

\begin{thebibliography}{10}

\bibitem{abnar2020quantifying}
Samira Abnar and Willem Zuidema.
\newblock Quantifying attention flow in transformers.
\newblock {\em arXiv preprint arXiv:2005.00928}, 2020.

\bibitem{brendel2019approximating}
Wieland Brendel and Matthias Bethge.
\newblock Approximating cnns with bag-of-local-features models works
  surprisingly well on imagenet.
\newblock {\em arXiv preprint arXiv:1904.00760}, 2019.

\bibitem{chapelle2010efficient}
Olivier Chapelle and S~Sathiya Keerthi.
\newblock Efficient algorithms for ranking with svms.
\newblock {\em Information retrieval}, 13(3):201--215, 2010.

\bibitem{dosovitskiy2020image}
Alexey Dosovitskiy, Lucas Beyer, Alexander Kolesnikov, Dirk Weissenborn,
  Xiaohua Zhai, Thomas Unterthiner, Mostafa Dehghani, Matthias Minderer, Georg
  Heigold, Sylvain Gelly, et~al.
\newblock An image is worth 16x16 words: Transformers for image recognition at
  scale.
\newblock {\em arXiv preprint arXiv:2010.11929}, 2020.

\bibitem{dubey2016deep}
Abhimanyu Dubey, Nikhil Naik, Devi Parikh, Ramesh Raskar, and C{\'e}sar~A
  Hidalgo.
\newblock Deep learning the city: Quantifying urban perception at a global
  scale.
\newblock In {\em European conference on computer vision}, pages 196--212.
  Springer, 2016.

\bibitem{geirhos2018imagenet}
Robert Geirhos, Patricia Rubisch, Claudio Michaelis, Matthias Bethge, Felix~A
  Wichmann, and Wieland Brendel.
\newblock Imagenet-trained cnns are biased towards texture; increasing shape
  bias improves accuracy and robustness.
\newblock {\em arXiv preprint arXiv:1811.12231}, 2018.

\bibitem{he2016deep}
Kaiming He, Xiangyu Zhang, Shaoqing Ren, and Jian Sun.
\newblock Deep residual learning for image recognition.
\newblock In {\em Proceedings of the IEEE conference on computer vision and
  pattern recognition}, pages 770--778, 2016.

\bibitem{krizhevsky2012imagenet}
Alex Krizhevsky, Ilya Sutskever, and Geoffrey~E Hinton.
\newblock Imagenet classification with deep convolutional neural networks.
\newblock {\em Advances in neural information processing systems},
  25:1097--1105, 2012.

\bibitem{muhammad2021eigen}
Mohammed~Bany Muhammad and Mohammed Yeasin.
\newblock Eigen-cam: Visual explanations for deep convolutional neural
  networks.
\newblock {\em SN Computer Science}, 2(1):1--14, 2021.

\bibitem{ramaswamy2020ablation}
Harish~Guruprasad Ramaswamy et~al.
\newblock Ablation-cam: Visual explanations for deep convolutional network via
  gradient-free localization.
\newblock In {\em Proceedings of the IEEE/CVF Winter Conference on Applications
  of Computer Vision}, pages 983--991, 2020.

\bibitem{ramirez2021measuring}
Tom{\'a}s Ram{\'\i}rez, Ricardo Hurtubia, Hans Lobel, and Tom{\'a}s Rossetti.
\newblock Measuring heterogeneous perception of urban space with massive data
  and machine learning: An application to safety.
\newblock {\em Landscape and Urban Planning}, 208:104002, 2021.

\bibitem{rossetti2019explaining}
Tom{\'a}s Rossetti, Hans Lobel, V{\'\i}ctor Rocco, and Ricardo Hurtubia.
\newblock Explaining subjective perceptions of public spaces as a function of
  the built environment: A massive data approach.
\newblock {\em Landscape and urban planning}, 181:169--178, 2019.

\bibitem{rudin2019stop}
Cynthia Rudin.
\newblock Stop explaining black box machine learning models for high stakes
  decisions and use interpretable models instead.
\newblock {\em Nature Machine Intelligence}, 1(5):206--215, 2019.

\bibitem{selvaraju2017grad}
Ramprasaath~R Selvaraju, Michael Cogswell, Abhishek Das, Ramakrishna Vedantam,
  Devi Parikh, and Dhruv Batra.
\newblock Grad-cam: Visual explanations from deep networks via gradient-based
  localization.
\newblock In {\em Proceedings of the IEEE international conference on computer
  vision}, pages 618--626, 2017.

\bibitem{shi2019deep}
Xiangwei Shi, Seyran Khademi, and Jan van Gemert.
\newblock Deep visual city recognition visualization.
\newblock In {\em CVPR Workshops}, pages 57--62, 2019.

\bibitem{sokol2020explainability}
Kacper Sokol and Peter Flach.
\newblock Explainability fact sheets: a framework for systematic assessment of
  explainable approaches.
\newblock In {\em Proceedings of the 2020 Conference on Fairness,
  Accountability, and Transparency}, pages 56--67, 2020.

\bibitem{touvron2021training}
Hugo Touvron, Matthieu Cord, Matthijs Douze, Francisco Massa, Alexandre
  Sablayrolles, and Herv{\'e} J{\'e}gou.
\newblock Training data-efficient image transformers \& distillation through
  attention.
\newblock In {\em International Conference on Machine Learning}, pages
  10347--10357. PMLR, 2021.

\bibitem{vidal2021deep}
Andr{\'e}s~C{\'a}diz Vidal.
\newblock {\em Deep Neural Network Models with Explainable Components for Urban
  Space Perception}.
\newblock PhD thesis, Pontificia Universidad Catolica de Chile (Chile), 2021.

\bibitem{viering2019manipulate}
Tom Viering, Ziqi Wang, Marco Loog, and Elmar Eisemann.
\newblock How to manipulate cnns to make them lie: the gradcam case.
\newblock {\em arXiv preprint arXiv:1907.10901}, 2019.

\bibitem{yao2019human}
Yao Yao, Zhaotang Liang, Zehao Yuan, Penghua Liu, Yongpan Bie, Jinbao Zhang,
  Ruoyu Wang, Jiale Wang, and Qingfeng Guan.
\newblock A human-machine adversarial scoring framework for urban perception
  assessment using street-view images.
\newblock {\em International Journal of Geographical Information Science},
  33(12):2363--2384, 2019.

\bibitem{zhang2018measuring}
Fan Zhang, Bolei Zhou, Liu Liu, Yu~Liu, Helene~H Fung, Hui Lin, and Carlo
  Ratti.
\newblock Measuring human perceptions of a large-scale urban region using
  machine learning.
\newblock {\em Landscape and Urban Planning}, 180:148--160, 2018.

\end{thebibliography}

\end{document}